%% file: acl_latex.tex
\pdfoutput=1

\documentclass[11pt]{article}

\usepackage[final]{acl}

\usepackage{times}
\usepackage{latexsym}
\usepackage{graphicx}
\usepackage{multirow}
\usepackage{multicol}
\usepackage{longtable}
\usepackage{xcolor}
\usepackage{float}
\usepackage[inline]{enumitem}

\newcommand{\method}[1][inline]{\textsl{SPARK}}

\usepackage[T1]{fontenc}

\usepackage[utf8]{inputenc}

\usepackage{microtype}

\title{Contextualizing Argument Quality Assessment with Relevant Knowledge}

\author{Darshan Deshpande$^1$,\qquad
Zhivar Sourati$^{1}$,\qquad
Filip Ilievski$^{2}$, \qquad Fred Morstatter$^1$ \\ 
$^1$Information Sciences Institute, University of Southern California\\
$^2$Department of Computer Science, Vrije Universiteit Amsterdam\\
  \texttt{\{darshang, souratih, fredmors\}@isi.edu,} \texttt{f.ilievski@vu.nl}}

\begin{document}
\maketitle
\begin{abstract}
Automatic assessment of the quality of arguments has been recognized as a challenging task with significant implications for misinformation and targeted speech.
While real-world arguments are tightly anchored in context, existing computational methods analyze their quality in isolation, which affects their accuracy and generalizability. 
We propose \method: a novel method for scoring argument quality based on contextualization via relevant knowledge. 
We devise four augmentations that leverage large language models to provide feedback, infer hidden assumptions, supply a similar-quality argument, or give a counter-argument.
\method~uses a dual-encoder Transformer architecture to enable the original argument and its augmentation to be considered jointly.
Our experiments in both in-domain and zero-shot setups show that \method~consistently outperforms existing techniques across multiple metrics. 

\end{abstract}

\input{sec/intro}
\input{sec/background}
\input{sec/enrichment}
\input{sec/experiments}
\input{sec/conclusion}
\input{sec/limitations}
\input{sec/ethics}
\bibliography{acl_latex}
\bibliographystyle{acl_natbib}


\appendix

\section{Appendix}
\label{sec:appendix}

\begin{table*}[!t]

\centering
\small
\begin{tabular}{|p{0.1\textwidth}|p{0.85\textwidth}|}
\hline

Feedback & - The argument takes a clear position that cell phones cause distraction while driving. \newline
- The writer expresses strong emotions regarding the topic, which may influence the persuasiveness of the argument.\newline
- The argument does not provide any evidence or supporting details beyond the writer's personal experience and emotions. \newline
- The argument does not address whether cell phones cause distractions in other contexts beyond driving.\\ \hline

Assumptions & - Cell phones cause a distraction when driving. \newline
- Bluetooth or speaker use of cell phones can also be a distraction while driving. \newline
- People don't take driving as seriously as they should. \\ \hline

Similar quality argument & Yes, cell phones can be a major distraction to people. People are constantly checking their phones for notifications, messages, and other tasks that can take away from what they should be focusing on. This can lead to people not paying attention in class, not being productive at work, and not being present in conversations. Cell phones are an amazing invention, but they can also be a major distraction.\\ \hline

Counter-Argument & 
While it is true that distracted driving can be a major problem, it is unfair to blame cell phones as the sole cause of distraction. Many other factors can also contribute to distracted driving, such as eating, applying makeup, or even talking with passengers. Furthermore, cell phones can also be used in a responsible manner while driving, such as with the use of hands-free devices or by pulling over to take a call. Therefore, instead of demonizing cell phones, we should focus on educating drivers on safe driving practices and the dangers of distracted driving in general.\\ \hline

\end{tabular}
\caption{Augmentation examples for the topic \textit{Do cell phones cause a distraction to people?} with argument: \textit{If you mean do they cause a distraction when driving the answer is a resounding YES!!! That includes bluetooth or speaker. Nothing makes me more angry when people don't take driving as seriously as they should. It is a major killer and if you don't follow the rules and focus you are putting other people in danger}.}
\label{tab:augmentation-examples}

\end{table*}
\begin{table*}[!t]

\centering
\small

\begin{tabular}{|p{0.1\textwidth}|p{0.85\textwidth}|}
\hline
Effectiveness Score & Generated similar quality argument to: Do you equate legality with morality? \\ \hline
1.0 & Yes, I think that if the law says something is wrong, then it is wrong. People should follow the law and not do anything illegal.\\ \hline
2.0 & No, I do not think that legality necessarily equates to morality. Laws can often be outdated, biased, or simply not reflective of what is right and wrong. For example, slavery used to be legal in many countries, yet it was obviously immoral. So legality and morality do not always align. \\ \hline

3.0 & No, I do not equate legality with morality. Just because something is legal does not make it moral, and just because something is illegal does not make it immoral. Laws are made by people and can be changed, while morality is often determined by an individual's personal values and beliefs. Therefore, legality should not be used as a measure of morality.\\ \hline

4.0 & 
No, I do not equate legality with morality. I believe that there are laws that should be followed in order to maintain the peace and order of a society, but there are also moral standards that must be adhered to for the same purpose. For example, it may be legal in some places to drive over the speed limit, but it is still morally wrong to do so. \\ \hline

5.0 & 
I do not necessarily equate legality with morality. Legality is based on laws, which can be changed to reflect the morality of a society. Morality, in contrast, is based on principles and values that don't necessarily have to be enforced by law. For example, while it may be legal to drive over the speed limit, it is not necessarily moral to do so. \\ \hline

\end{tabular}
\caption{Similar quality examples.}
\label{tab:similar-quality-examples}

\end{table*}

\subsection{GPT 3.5 prompt templates}

\textbf{Feedback}\\ 
The feedback on writing is sampled by considering both the topic and the argument related to the topic. To ensure brevity, we output the feedback in bullet point format. We follow the format below for sampling the feedback from the LLM:\newline\newline\textit{Give concise writing feedback for the following argument in context with the topic, preferably in bullet points}:\newline \textit{Topic}: \underline{\textit{{topic}}} \newline \textit{Argument}: \underline{\textit{{argument}}}.
\newline\newline
\textbf{Assumptions}\\
Similar to feedback, assumptions are sampled in bullet point format to ensure brevity. Additionally, to constrain the hallucinations of the LLM, we restrict it to output "No assumptions" for the cases where it does not find assumptions or biases. We use the below prompt to sample this assumptions list:\newline \newline
\textit{Summarize the assumptions, if any, in the following argument in a bullet format otherwise return "No assumptions" \newline Topic: \underline{\textit{{topic}}} \newline Argument: \underline{\textit{argument}}}.\newline \newline
\textbf{Similar quality argument}\newline
To sample a similar quality argument, we use the following template:\newline\newline
\textit{Cogency Score:}\underline{\textit{ {cogency score}}}\newline
\textit{Effectiveness Score:} \underline{\textit{{effectiveness score}}} \newline
\textit{Reasonableness Score: }\underline{\textit{{reasonableness score}}} \newline
\textit{Topic:} \underline{\textit{{topic}}} \newline\newline
We use ten samples in the few shot setting with two each from every integer ranking from 1-5 on the ranking scale for each metric.
Finally, we prompt the LLM to generate the argument with respect to the cogency, effectiveness and robustness scores.\newline\newline
\textbf{Counter-argument}\\
The counter-argument is generated using the given argument and topic, and the following template:\newline\newline
\textit{Give a counter-argument for the following argument with respect to the Topic: \underline{\textit{{topic}}} \newline Argument: \underline{\textit{{argument}}}} \newline

\begin{figure*}
\centering
\begin{minipage}[b]{0.3\textwidth}
  \centering
  \includegraphics[width=\textwidth]{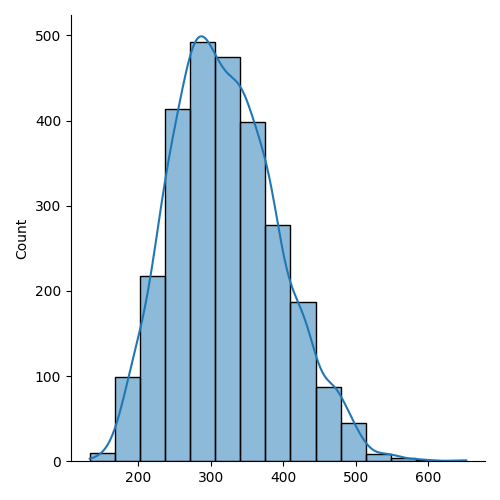}
\end{minipage}
\hfill
\begin{minipage}[b]{0.3\textwidth}
  \centering
  \includegraphics[width=\textwidth]{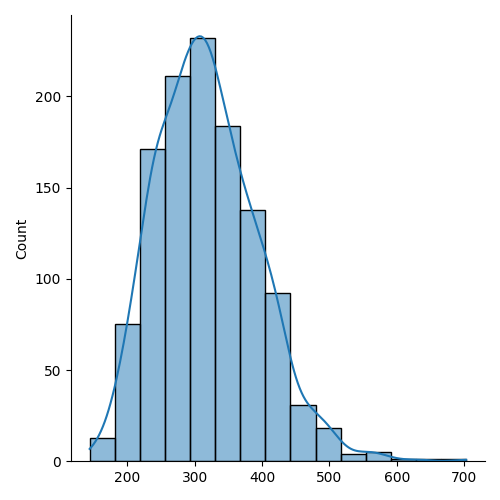}
\end{minipage}
\hfill
\begin{minipage}[b]{0.3\textwidth}
  \centering
  \includegraphics[width=\textwidth]{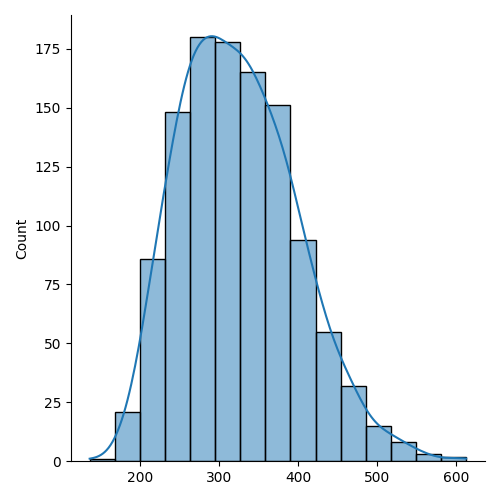}
\end{minipage}
\caption{Distributions for augmentation lengths for the training, validation, and testing splits, respectively.}
\label{fig:distributions}
\end{figure*}

\begin{table*}[]
    \centering
    \begin{tabular}{|p{0.15\textwidth}|p{0.15\textwidth}|p{0.1\textwidth}|p{0.1\textwidth}|p{0.1\textwidth}|p{0.1\textwidth}|}
    \hline
         Dataset split & Dataset size & Minimum length & Maximum length & Mean length & Variance \\
         \hline
         Train & 2746 & 133 & 652 & 320.27 & 5309.61 \\
         Validation & 1177 & 144 & 703 & 318.55 & 5690.65 \\
         Test & 1139 & 136 & 613 & 324.60 & 5467.65 \\ \hline
    \end{tabular}
    \caption{Distribution properties across splits for the concatenated augmentations.}
    \label{tab:distribution_stats}
\end{table*}

\subsection{Augmentation examples}
\label{appendix-aug-examples}

\autoref{tab:augmentation-examples} shows augmentation examples for one topic-argument pair. The topic questions if cell phones distract people and the argument agrees with it in context to distracted driving due to cell phones.

\textbf{Feedback: }
The feedback discusses how the argument takes a clear position but appears overly emotional while answering, which may influence the persuasiveness of the argument. Additionally, the feedback regarding lack of evidence other than personal experience and the lack of discussion on phones causing a distraction beyond driving aim to help improve understanding while ensuring that the model pays careful attention to the topic in context.

\textbf{Assumptions:}
The LLM lists several important assumptions in the proposed argument with respect to the topic. The first is the base assumption of the author's perspective, which is that cell phones cause distraction during driving.  The LLM extracts the sentence \textit{" people don’t take driving as seriously as they should"} and labels it as an assumption because this is a faulty generalization to apply a general rule to all people, which is, in this case, people not taking driving seriously. 

\textbf{Similar quality argument: }
The generated similar quality argument tries to replicate the structural pattern of the given argument. The similar quality instance contains open-ended, long-winded sentences such as \textit{"Cell phones are an
amazing invention, but they can also be a major distraction"} which reduce the score of the argument, similar to the original argument. We also see that the LLM understands the ranking progression in a few shot setting and similar to the original argument, the similar quality argument also focuses solely on driving based cell phone distractions.

\textbf{Counter argument:}
We notice that the LLM recognizes that the original argument only discusses distracted driving and so it only produces a counter argument of the stance that distracted driving is not the only cause for distraction. The response discusses the safe use of cell phones such as hands free, etc and advocates educating drivers on the effects of cell phone based distracted driving.

\subsection{Impact of effectiveness score on GPT 3.5 outputs for similar quality arguments}

Table ~\ref{tab:similar-quality-examples} discusses the impact of effectiveness score on the generated similar quality argument. As can be seen in Table \ref{tab:similar-quality-examples}, the generated argument with an effectiveness score of 1.0 oversimplifies the relationship between legality and morality and treats related laws as fixed. Comparatively, the argument with an effectiveness score of 2.0 provides an example of slavery which enhances the effectiveness of the argument. Despite the addition of this example, the second argument lacks elaboration on why the example is immoral and fails to provide relevant evidence. The argument generated, given an effectiveness score of 3.0, recognizes that the law is not the sole arbiter of morality and that laws are subject to change. It does not only highlight the potential flaws in legal systems but also addresses the distinction between personal values and the law. However, this argument oversimplifies morality by implying that personal values and beliefs solely determine morality and lacks supporting evidence for the statement: \textit{just because something is legal does not make it moral, and just because something is illegal does not make it immoral.}. The argument generated with an effectiveness score of 4.0 considers the coexistence of legal and moral standards. The addition of a specific example in this argument adds concreteness and strengthens its persuasiveness. However, the argument can be further strengthened by acknowledging a broader range of situations where legality and morality may diverge.
Finally, the argument ranked with the highest effectiveness score emphasizes the independence of morality from legal enforcement, which makes it even more persuasive. The contrasting comparison adds clarity to the flow of the argument and hence makes it better than all previously generated arguments.

\subsection{Distribution analysis of augmentation lengths across splits} 
\label{augmentation_distribution_analysis}

In this subsection, we conduct a distribution analysis on the augmentation input sizes to justify the use of the dual BERT architecture. Based on the findings presented in \autoref{tab:distribution_stats}, the training split exhibits a minimum tokenized sequence length of 133 tokens, a maximum length of 652 tokens, and an average length of 320.27. The distribution of the training set, as presented by \autoref{fig:distributions}, shows that only 0.5\% (15 examples) of the training data exceeds BERT's token limit of 512 tokens.

The validation split has a minimum tokenized length of 144 tokens, a maximum length of 703 tokens, and an average length of 318.55 tokens. The testing split, on the other hand, has a minimum length of 136 tokens, a maximum length of 613 tokens, and an average length of 324.60 tokens. The percentage of data points in the validation and testing splits as seen in \autoref{fig:distributions} that exceed BERT's token limit are only 1.1\% (13 examples) and 1.31\% (15 examples), respectively.

Hence, we can conclude that the second BERT encoder tasked with embedding the augmentations is able to capture all the information in the augmentations without truncating the augmentations.



\subsection{Implementation and Human Evaluation Specifications}
\subsubsection{Training setup}
We used the Hugging Face library~\cite{wolf2020huggingfaces} for the training and inference of our models. The training of the dual BERT architecture was performed using 6xA5000 GPUs; each training run for five epochs with a batch size of 32 on each device and a learning rate of $5 \times 10^{-3}$ takes approximately 30 minutes to converge. We found that the convergence was sensitive to the learning rate and smaller learning rate within the range of $1 \times 10^{-4}$ to $5 \times 10^{-3}$ was preferable for optimal results. The cosine scheduler provided the fastest convergence out of the possible linear, cosine, and polynomial scheduler options, and the best model was picked according to the highest F1 score on the validation split of the GAQCorpus. The inference for Llama-2 (7B) and Flan T5-XL (2.85B) was performed on 6xA5000 GPUs and took approximately 7 hours to complete.
The GPT-3.5 experiments were repeated three times, and the average of the three runs is reported in~\autoref{tab:results}. We maintained a temperature and top\_p of 0.01 and 0.9, respectively, for all LLM experiments.

\subsubsection{Human Evaluation}
Our human study poses the following three targeted questions to the participants:
\begin{enumerate}
    \item How \textbf{valid} is the information provided by the augmentation with respect to the background of the argument?
    \item How \textbf{informative} is the augmentation for the task of argument quality analysis?
    \item How \textbf{relevant} is the augmentation to help with the task of assessing the quality of the argument?
\end{enumerate}

Our study was done with a sample of five computer science graduate students at the University of Southern California, and the results were averaged and reported in \autoref{tab:user_study}. To mitigate potential issues arising from sensitive arguments, we ensured that the questions were clearly understood and terms were explained, after which we obtained oral consent from each participant. The study was exempted from review by IRB (application UP-21-00443).

\end{document}

%% file: sec/intro.tex
\section{Introduction}
Reliable analysis of arguments in natural language holds the promise to support applications such as automated grading \cite{automated_essay_scoring}, and tackling misinformation and targeted speech \cite{alhindi_misinformation_2023}. 
Computational argument analysis has been relatively popular through tasks like
argument extraction~\cite{arg_extraction_ampersand},
evidence mining~\cite{evidence_extraction_rinott-etal-2015-show}, relation assignment \cite{relational_models_Trautmann2020}, writing support \cite{writing_support_stab-gurevych-2014-identifying} and claim generation \cite{bilu-slonim-2016-claim_generation}. 
A particularly challenging task is argument quality assessment \cite{fromm2022holistic}, which addresses the cogency, effectiveness, and reasonableness of an argument~\cite{wachsmuth-etal-2017-computational} pertaining to a topic.
Assessing the quality of the argument involves analyzing the objective evidence, relevant assumptions, and structural soundness, making the overall task difficult.

\begin{figure}
    \includegraphics[width=0.5\textwidth]{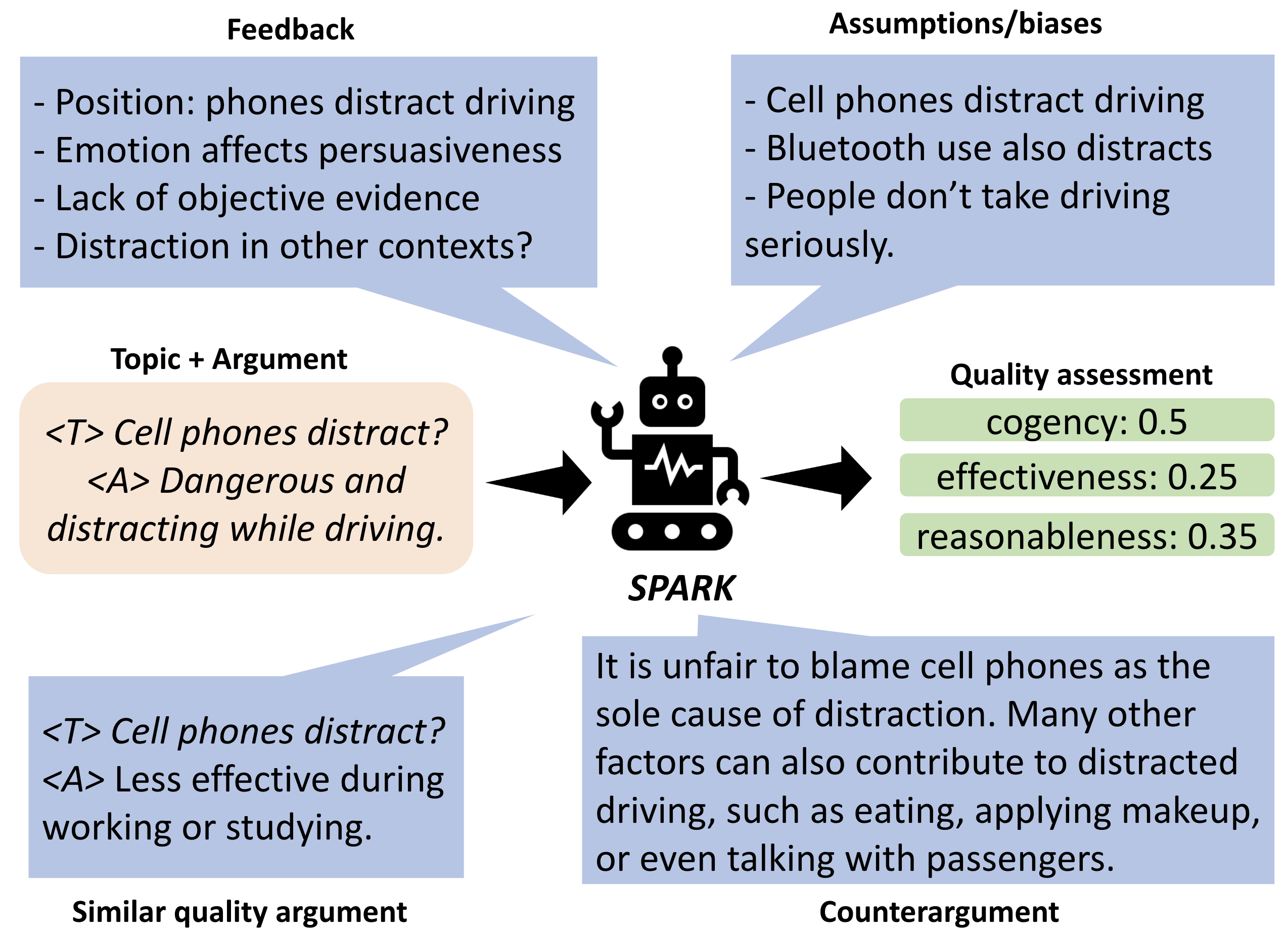}
    \caption{Overview of \method.}
    \label{fig:overview}
    \vspace{-2em}
\end{figure}

Research on argument quality assessment has focused on extracting textual patterns using various learning frameworks and content features~\cite{scientia_potentia_lauscher_etal}. It has been widely recognized that contextualizing arguments with implicit knowledge, such as extracting claim revisions~\cite{skitalinskaya2021learning} and generating explicit conclusions~\cite{gurcke2021assessing} can be informative for reasoning models. However, we note that: 1) these methods fail to generalize to novel arguments where this information is not available, and 2) no prior work has considered jointly a comprehensive set of such contextualization strategies. 

To bridge these gaps, we propose a novel framework called \method~(\textbf{S}coring the \textbf{P}ragmatics of \textbf{A}rguments via \textbf{R}elevant \textbf{K}nowledge), which 
incorporates augmentation strategies based on a large language model
(LLM), GPT 3.5 \cite{chatgpt}, and elements from argumentation literature~\cite{nickerson_2020,Mulyati2023,harvey2009brief}, specifically, feedback, assumptions, arguments with similar quality, and counter-arguments. \method~ processes the original argument and topic and its augmentations separately using a dual-encoder Transformer architecture with a multi-head cross-attention layer. 
We demonstrate the effectiveness of \method's augmentations and architecture using both in-domain and out-of-domain datasets. 
We make our entire code available at 
\url{https://github.com/usc-isi-i2/forecast-argument}.

%% file: sec/background.tex
\vspace{-0.4em}
\section{Background}
\label{ssec:task_formulation}
\textbf{Task formulation.}
Inspired by prior work ~\cite{ibm30k_Gretz_Friedman_CohenKarlik_Toledo_Lahav_Aharonov_Slonim_2020,lauscher-etal-2020-rhetoric}, we formalize argumentation quality assessment as a regression task of predicting the quality of a natural language argument. Given a topic and an argument, we consider three quality indicators~\cite{lauscher-etal-2020-rhetoric}: 1) \textit{cogency}, which evaluates the relevance and sufficiency of the argument’s premise in relation to the conclusion, 2) \textit{effectiveness}, which measures the argument’s persuasive power based on factors like arrangement, clarity, and appropriateness, and 3) \textit{reasonableness}, which determines the argument’s ability to resolve the debate’s issue~\cite{wachsmuth-etal-2017-argumentation}. The overall quality of an argument can be estimated by averaging these three metrics~\cite{ibm30k_Gretz_Friedman_CohenKarlik_Toledo_Lahav_Aharonov_Slonim_2020}.

\noindent\textbf{Connection to prior studies.}
The introduction of benchmarks for argument quality~\cite{stab-gurevych-2014-annotating,ibm30k_Gretz_Friedman_CohenKarlik_Toledo_Lahav_Aharonov_Slonim_2020} has inspired various methods based on logistic regression~\cite{ghosh-etal-2016-coarse}, fully connected and recurrent neural networks~\cite{habernal-gurevych-2016-makes}, and fine-tuned Transformers~\cite{toledo-etal-2019-automatic}.
\citet{hulpus2019towards} explained that contextualizing an argument with implicit knowledge is essential to understanding its quality.
\citet{scientia_potentia_lauscher_etal} categorized knowledge used by current argument assessment research so far under linguistic, task-specific, and argument-specific sections.
To mimic human reasoning over arguments, prior work has incorporated users' prior beliefs as predictors of argument persuasiveness~\cite{durmus-cardie-2018-exploring}, trained classifiers for different audience groups \cite{el-baff-etal-2020-analyzing}, utilized user history to predict persuasion \cite{al-khatib-etal-2020-exploiting}, augmented arguments with supporting or refuting documents~\cite{marro-etal-2022-graph}, and augmented arguments with visual features~\cite{hasan-etal-2021-hitting}. 
Most similar to \method, \citet{skitalinskaya2021learning} leverage comparison between revisions of the same claims, while \citet{gurcke2021assessing} generate conclusions to assess argument sufficiency. However, revisions are rarely provided for novel arguments, whereas generated conclusions are argument-specific and may not generalize well~\cite{gurcke2021assessing}.
Addressing prior work limitations, \method~implements four well-motivated augmentation strategies to enhance novel arguments and utilizes an attention-based dual encoder model for effective reasoning.

%% file: sec/enrichment.tex
\vspace{-0.4em}
\section{\method}


\paragraph{Augmentation strategies.} We devise four augmentation techniques to contextualize arguments. We generate the augmentations by prompting
GPT-3.5 \cite{chatgpt} (see appendix for details).

\noindent \underline{\textit{Feedback.}}
Constructive feedback in the form of comments and suggestions helps comprehension and domain knowledge acquisition~\cite{Mulyati2023}. We hypothesize that assessing argument strengths and weaknesses helps argument ranking, as in \autoref{fig:overview}, where feedback identifies emotional appeal, insufficient evidence, and generalization. We prompt the LLM to generate writing feedback for a topic-argument pair in a zero-shot setting.

\noindent \underline{\textit{Assumptions.}}
Unstated assumptions frequently introduce bias in arguments~\cite{nickerson_2020}. Making assumptions explicit can reveal these hidden biases, which may aid in assessing argument persuasiveness and relevance. One such assumption in \autoref{fig:overview} is that people do not take driving seriously. We employ an LLM to extract the argument's underlying assumptions in a zero-shot setting.

\noindent \underline{\textit{Similar-quality instance.}}
Inspired by prior work on claim revisions~\cite{skitalinskaya2021learning}, 
we hypothesize that retrieving arguments with similar quality at training time leads to generalizable model learning.
For this purpose, we derive a synthetic argument with similar reasonableness, cogency, and effectiveness to the original one (\autoref{fig:overview}). 
We generate this synthetic argument 
in a few-shot setting, where the LLM has access to example arguments alongside their quality scores covering the full 1-5 range. Since this augmentation uses ground-truth information that is not available during inference, we randomly replace synthetic arguments with \textit{None} at training time with a probability of $P=0.5$, thus familiarizing the model with the absence of similar arguments during testing where it only sees \textit{None}.
This technique is similar to distillation~\cite{hinton2015distilling}, where the encoder's (student) goal is to utilize the LLM's (teacher) ranking-based argument generations (soft labels) to learn the argument quality ranking task.

\noindent \underline{\textit{Counter-arguments.}}
Counter-arguments provide objections, alternatives, and doubts of a skeptical reader \cite{harvey2009brief}.
We expect that contrasting the strengths and weaknesses of two opposite arguments will aid quality assessment.
The example counter-argument in~\autoref{fig:overview} makes a firm claim that alternative activities (e.g., eating) lead to distractions, and solely blaming phones is unfair.
We ask the LLM to provide a counter-argument for a topic and an argument in a zero-shot setting. 

\begin{figure}
\centering
\includegraphics[width=\linewidth]{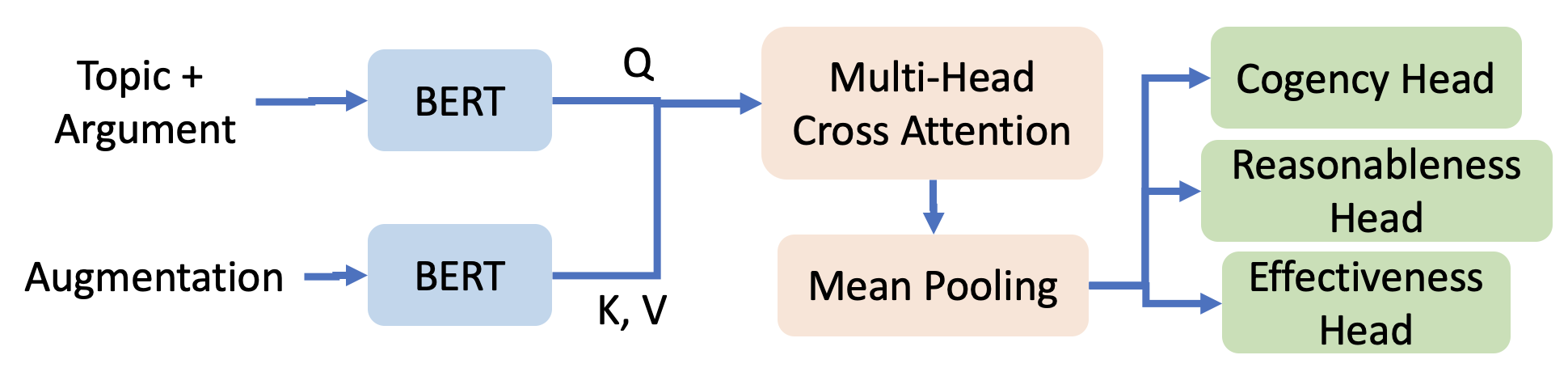}
  \caption{Dual BERT encoder architecture.}
  \label{fig_architecture}
  \vspace{-1em}
\end{figure}
\paragraph{Dual-encoder architecture.} To consider the argument together with the augmentations, we employ a dual BERT encoder (\autoref{fig_architecture}) as an improvement to the architecture by \citet{dual_end_to_end_retrieval}.
The first encoder embeds the topic and argument, whereas the second embeds the augmentations.
The second encoder can store individual augmentations or their concatenation, arbitrarily fixed to 
\textit{Similar quality argument [SEP] Feedback [SEP] Assumptions [SEP] Counter-argument}.
Notably, the dual encoder can effectively store all of the augmentation data without truncating information in practically all cases (see \autoref{augmentation_distribution_analysis} for the augmentation lengths). 
We use a multi-head cross-attention layer \cite{attention_is_all_you_need} to enable the model to weigh each augmentation according to the argument-topic pair. We pass the attention outputs to a mean pooler, whose output is fed into three separate regressor heads, one per quality metric.

%% file: sec/experiments.tex
\begin{table*}[!ht]
\centering
\small

\begin{tabular}{|l l|c|c|c|c|c|c|c|c|}
\hline
& & \multicolumn{6}{|c|}{GAQ Corpus (in-domain)} & \multicolumn{2}{|p{2.1cm}|}{IBM-30K (ZS)} \\
\cline{3-10}
\textbf{Model} & \textbf{Augmentation} & \multicolumn{2}{c|}{Cogency} & \multicolumn{2}{c|}{Effectiveness} & \multicolumn{2}{c|}{Reasonableness} & \multicolumn{2}{c|}{WA} \\
\cline{3-10}

 & & $\sigma$ & $\rho$ & $\sigma$ & $\rho$ & $\sigma$ & $\rho$ & $\sigma$ & $\rho$ \\ \hline

BERT & - & 0.3480 & 0.3268 & 0.2804 & 0.2821 & 0.3285 & 0.3356 & 0.1989& 0.1751\\

XLNet & - & 0.1790 & 0.1673 & 0.2008 & 0.1930 & 0.1778 & 0.1847 & 0.1989 & 0.1816\\ 

Dual BERT & - & 0.3685 & 0.3619 & 0.3082 & 0.3143 & 0.3694 & 0.3848 &  0.1766 & 0.1588\\

GPT-3.5 & - & 0.2879 & 0.3146 & 0.3585 & 0.3902 & 0.3561 & 0.4073 & 0.2698 & 0.2794 \\
\hline

BERT & DPR & 0.3215 & 0.3182 & 0.2728 & 0.2763 & 0.2821 & 0.3012 & 0.0706 & 0.0774 \\

XLNet & DPR & 0.2459 & 0.1778 & 0.2259 & 0.2271 & 0.2024 & 0.2160 & 0.1201 & 0.1254 \\
Dual BERT & DPR & 0.3536 & 0.3525 & 0.3227 & 0.3224 & 0.3396 & 0.3497 & 0.1571 & 0.1408\\ \hline
Dual BERT &  Flan T5 XL (all) & 0.3262 & 0.3268 & 0.2850 & 0.2920 & 0.3445 & 0.3564 & 0.1034 & 0.1241 \\
Dual BERT & Llama-2 (7B) (all) & 0.3468 & 0.3516 & 0.3296 & 	
0.3516 & 0.3269 & 0.3529 & 0.1673 & 0.1398 \\
\hline

BERT & GPT-3.5 (all) & 0.3418 & 0.3340 & 0.2863 & 0.3047 & 0.3459 & 0.3664 & 0.1510 & 0.1374\\
XLNet &  GPT-3.5 (all) & 0.2675 & 0.1930 & 0.2450 & 0.2060 & 0.2142 & 0.2046 & 0.1318 & 0.1352 \\ 
GPT-3.5 & GPT-3.5 (all) & 0.3491 & 0.3455 & 0.3660 & 0.4050 & 0.3596 & 0.3974 & \underline{0.2710} & \underline{0.2804}\\
Dual BERT & GPT-3.5 (feedback)  & \underline{0.4184} & \underline{0.4187} & 0.3925 & 0.4037 & \textbf{0.4174} & \underline{0.4318} & \textbf{0.2742} & \textbf{0.2854}\\ 
Dual BERT & GPT-3.5 (assumptions)  & 0.3699 & 0.3704 & 0.3881 & 0.3946 & 0.3681 & 0.3806 & 0.1728 & 0.1769 \\ 
Dual BERT & GPT-3.5 (similar quality) & 0.4059 & 0.4038 & \underline{0.4150} & \underline{0.4252} & 0.3545 & 0.3775 & 0.2629 & 0.2037 \\
Dual BERT & GPT-3.5 (counter)  & 0.4030 & 0.4092 & 0.4048 & 0.4143 & 0.3989 & 0.4171 & 0.2086 & 0.2044  \\ 
\textbf{Dual BERT} & \textbf{GPT-3.5 (all)} & \textbf{0.4242} & \textbf{0.4371} & \textbf{0.4513} & \textbf{0.4762} & \underline{0.4135} & \textbf{0.4362}  & 0.2238 & 0.2293 \\

\hline
\end{tabular}
\caption{Performance of Dual-BERT model with augmentations applied compared to the baseline models. The performance of the model achieving the best scores per metric is \textbf{boldfaced} and the second best score is \underline{underlined}.}
\label{tab:results}

\end{table*}
\begin{table}[!ht]
    \centering
    \small
    \begin{tabular}{|p{0.16\textwidth}|r|r|r|}
    \hline
         \textbf{Aug. / Metric} & \textbf{Validity} & \textbf{Inform.} & \textbf{Relevance} \\ \hline
         Feedback & 4.87 & 3.80  & 4.96 \\
         Assumptions & 4.95 & 4.91 & 4.97 \\
         Counter-arguments & 4.93 & 4.94 & 4.99 \\
         Similar quality & 4.40  & 4.40  & 4.62 \\\hline
    \end{tabular}
    \caption{Validity, informativeness, and relevance scores of the augmentations for argument scoring.}
    \label{tab:user_study}
\end{table}

\section{Experiments}
\subsection{Baselines}

\paragraph{Scoring models.}
We compare our dual BERT with a standard BERT model \cite{devlin-etal-2019-bert} to contrast the effect of disjoint embeddings against a concatenated input similar to the one used for dual BERT.
We compare dual BERT to  XLNet \cite{xlnet}, as they can both handle more than 512 tokens.
Finally, we utilize GPT-3.5 in a zero-shot setting
to gauge the model's ability to accomplish the task directly, without a dual encoder. We provide GPT-3.5 with the definitions of each metric and prompt it to individually rate each argument by a float between 1 and 5
with respect to the topic.

\vspace{-0.65em}
\paragraph{Alternative augmentation strategies.}
We evaluate the impact of using all augmentations together or one at a time, against ablated baselines without augmentations. We include two alternative augmentation methods: 
Wikipedia paragraphs extracted using dense passage retrieval (DPR)~\cite{karpukhin-etal-2020-dense-passage-retrieval}, and augmentations generated using smaller models, namely, Flan-T5-XL~\cite{flan-t5} and Llama-2 (7B)~\cite{touvron2023llama}.
\vspace{-0.3em}
\subsection{Datasets and Evaluation}
We use \textit{GAQCorpus} \cite{lauscher-etal-2020-rhetoric} as our training dataset for its diversity of domains (reviews, QA, and debates) and quality metrics (cogency, effectiveness, and reasonableness). 
We also use \textit{IBM-30K} \cite{ibm30k_Gretz_Friedman_CohenKarlik_Toledo_Lahav_Aharonov_Slonim_2020} to test \method's generalization on out-of-domain data. For IBM-30K, we perform weighted averaging (WA) of the three metric scores, which is supported by the high correlation between IBM-30K's WA and the GAQCorpus metrics \cite{lauscher-etal-2020-rhetoric}.
We report the Pearson ($\rho$) and Spearman ($\sigma$) correlation coefficients between the predictions and ground truth.

\subsection{Results}

We investigate \textit{whether augmentations help language models assess the quality of arguments more effectively} (Q1); \textit{how augmentation strategies compare to each other} (Q2); \textit{whether human quality judgments align with their model utility} (Q3) and \textit{how augmentations affect quality scores} (Q4).
\paragraph{Effect of augmentation on in-domain performance (Q1).}
\autoref{tab:results} shows that the overall best-performing combination uses Dual BERT with all four augmentations combined, which improves the Spearman correlation over the baseline BERT by 0.08-0.17 across the three metrics. The improvement is the largest for effectiveness, where the Spearman correlation increases by 61\%. While both single BERT, XLNet, and GPT-3.5 benefit from \method's augmentations as well, their performance is consistently lower than using the dual encoder. GPT-3 alone often performs better than the other baselines but lags significantly behind \method, showing the importance of the dual encoder.
Among the augmentations, the benefit of our four augmentations declines when using a smaller generative model (Flan T5 and Llama-2), augmenting via DPR, or using the dual BERT with a masked second encoder. Thus, merely adding text to the second encoder does not by itself bring higher performance.
The gap between \method~and the baselines increases in the zero-shot setting on the IBM-Rank-30K dataset. Here, augmentation with DPR, Flan-T5, and Llama-2 is consistently inferior, as is the \method~augmentation of the single-encoder methods. 
In summary, \method~effectively combines dual encoding and data augmentation for strong task accuracy and generalization.

\paragraph{Comparison of augmentation variants (Q2).} 
On the in-domain task, the dual encoder performs the best when it has access to the information from the four \method~ augmentations simultaneously. Among them, feedback is most effective for predicting argument cogency and reasonableness as it exposes flaws that directly relate to these metrics. 
Meanwhile, contextualizing through similar-quality arguments is optimal for predicting effectiveness, which we attribute to illustrating the connection between quality scores and the argument structure, format, and wording.
On out-of-domain data, the best performance is obtained by the feedback-augmented dual BERT, which even outperforms using all augmentations. 
This is illustrated in \autoref{tab:augmentation-examples}, where the first two arguments receive positive feedback, with space for improvement by further elaboration or addressing of alternatives, directing \method~ to increase its score. The third argument receives more critical feedback, causing \method~ to decrease its score. While GPT-3.5 is generally able to highlight the salient points of an argument and provide valid criticism, we also note an occasional bias of the model towards maintaining a neutral or positive argument stance (as in the case of the libertarianism argument).

\paragraph{Human judgment of augmentations (Q3).} 
To validate the alignment of the augmentations with human utility, we performed a human study where we asked participants to score the validity, informativeness, and relevance of each augmentation strategy.
The participants were asked to score 50 randomly sampled in-domain data points on these three metrics using a Likert scale of 1 (lowest) to 5 (highest).
The results in \autoref{tab:user_study} show that the augmentations are perceived by people as highly valid, informative, and relevant. Assumptions and counter-arguments were found to be consistently more valid, informative, and relevant than the other augmentations. Curiously, the participants judged the feedback informativeness to be lower, explaining that it often summarizes the argument instead of giving writing suggestions.
This finding provides a cue that highly effective augmentations for models may not be perceived as informative by people. 
\method~alleviates this issue by effectively combining the complementary augmentations and delegating the weighting of their utility to the model.

\begin{table*}[!ht]
    \centering
    \small
    \begin{tabular}{|p{0.22\textwidth}|p{0.10\textwidth}|p{0.28\textwidth}|p{0.28\textwidth}|}
    \hline
         Topic + Argument & Ground truth score & Feedback & Assumptions \\\hline
         Topic: The use of public defenders should be mandatory \newline \newline Argument: A centralized system of criminal defense would mean that all people would have access to the same standard of legal counsel, meaning that wealth and power can't be used to avoid justice. & 0.628 &  - Clear and concise argument presented 
         
         - Supports the idea of a centralized system of defense 
         
         - Advocates equal access to legal counsel
         
         - Addresses the issue of wealth and power influencing justice
         
         - Could benefit from further elaboration or evidence to strengthen the argument. ($0.602 \rightarrow 0.675$)& 
         
         - Assumes that a centralized system of criminal defense would provide the same standard of legal counsel to all individuals.
         
         - Assumes that wealth and power currently enable some individuals to avoid justice.
         
         - Assumes that making public defenders mandatory would address disparities in legal representation. ($0.602 \rightarrow 0.645$)\\\hline

        Topic: We should ban algorithmic trading \newline \newline Argument: algorithmic trading has been responsible for several mini market collapses, since computer systems lack the human sensitivity to look outside the stream of meaningless numbers to a wider context. & 0.948 & - Clear and concise argument 
        
    - Provides specific examples to support argument
    
    - Could benefit from further elaboration on the potential consequences of mini market collapses
    
    - Could also benefit from addressing potential counterarguments or alternative solutions to the issue. ($0.538 \rightarrow 0.688$)&
    - Algorithmic trading has been responsible for several mini market collapses.
    
    - Computer systems lack human sensitivity to look outside the stream of meaningless numbers.
    
    - Banning algorithmic trading will prevent or reduce mini market collapses. ($0.538 \rightarrow 0.5614$)\\
\hline

    Topic: We should adopt libertarianism 
    \newline \newline Argument: libertarianism is a justification for greed and exploitation &  0.666 &  The argument is not specific enough about what adopting libertarianism entails
    
    - It assumes that libertarianism automatically leads to working together for a greater good and favoring the less well off, which is not necessarily true
    
    - The argument could benefit from providing concrete examples of how libertarianism would benefit society
    
    - It is unclear how freedom of choice would lead to greater societal benefits. ($0.513 \rightarrow 0.585$)& 
    
    - Assumes that libertarianism is solely about justifying greed and exploitation
    
    - Assumes that there are no other principles or values within libertarianism
    
    - Assumes that all proponents of libertarianism advocate for greed and exploitation. ($0.513 \rightarrow 0.518$)\\\hline
    \end{tabular}    
    \caption{Feedback and assumption examples for the out-of-domain dataset. The model-predicted weighted-average scores with augmentation are compared against no-augmentation results in the parenthesis (w/o aug $\rightarrow$ w/aug) after the respective augmentations.}
    \label{tab:augmentation_ood}
\end{table*}

\paragraph{How augmentations affect quality scores (Q4)}
\autoref{tab:augmentation_ood} shows the impact on quality scores before and after augmentation generated by SPARK. Here, we contrast augmentation with feedback against assumptions (best-performing variants in the model and human evaluation) to study the reason for the low human informativeness scores for feedback augmentation. The generated assumptions are mostly understandable and accurate but not always exhaustive. We also note that while the model scores improve, it still cannot utilize assumptions as well as feedback for the ranking task. Meanwhile, feedback generation could identify weaknesses accurately; however, in some cases, as illustrated by the final example, the LLM misidentifies the stance of the argument for libertarianism and recommends that the participant justify how society can benefit from it. Such behavior occurs for sensitive socio-political topics and likely stems from the debiasing and safety measures adopted by OpenAI. Despite incorrect stances, the model produced better scores with augmentation, suggesting it recognizes patterns for ranking that differ from human reasoning for argument quality analysis.

%% file: sec/conclusion.tex
\vspace{-0.4em}
\section{Conclusions}
This paper enhances argument quality estimation models by providing contextualized feedback, inferred assumptions, similar quality arguments, and counter-arguments. We employ a dual-encoder Transformer to compare the argument and additional evidence effectively. Experimental results demonstrate that the best performance is achieved 
with the combination of all augmentations,
indicating their complementary insights. Our method, \method, outperforms single BERT, XLNet, and GPT-3.5, surpassing baselines and alternative augmentations in both in- and out-of-domain scenarios. Feedback augmentation is the most effective augmentation strategy for models, despite being scored the least informative by humans, because the model pattern matching differs from human reasoning.

%% file: sec/limitations.tex
\section{Limitations}

The limitations in this paper primarily stem from the generative capabilities and hallucination tendencies of the LLM used for augmentation. For example, despite constraining the output format for the assumption augmentation task, the LLM still generates "No assumptions" as output after listing a set of valid assumptions. As future work, further augmentation studies must be performed to analyze and improve the prompts to minimize the misunderstanding and biases of the LLM. LLMs should also be combined with other, possibly symbolic, components that can mitigate challenges with bias and hallucinations.

Moreover, evaluating in a larger set of domains, languages besides English, and other argumentative tasks such as logical fallacy detection is an important next step in investigating the generalizability of \method. While in theory, \method~can be directly applied to such tasks, it remains to be seen to which extent the current architecture and augmentation strategies will generalize to other tasks.

Finally, our analysis in this paper focuses on the comparison of methods and strategies, yet, we do not dive deep into the specific differences in performance across the three quality metrics, which serves as an important future direction.

%% file: sec/ethics.tex
\section{Ethics Statement}

Using LLMs like GPT-3.5, with large and inaccessible pretraining corpora, can potentially lead to the amplification of biases in downstream argument quality ranking models. While we believe that these biases are sometimes necessary to successfully judge the quality of the argument, sampling assumptions from such models can lead to biased and unethical information being fed to (and amplified by) the model. 
A possible way to minimize this harmful knowledge for scenarios that involve sampling assumptions is by providing a fallback instruction in the prompt for the model to output "No assumptions"; in the present paper, we did not conduct specific studies to measure the impact of this fallback strategy because we did not see a significant impact on a small number of samples. Furthermore, to be able to compare with pre-existing baselines, we did not de-bias or anonymize the datasets provided but we strongly suggest that this should be considered wherever the~\method~ method is deployed. 